%% file: paper.tex
\documentclass[onecolumn]{fairmeta}
\providecommand{\ours}{\textsc{TrustMem}\xspace}
\providecommand{\yes}

\usepackage{multirow}
\usepackage[table]{xcolor}
\usepackage{caption}
\usepackage{wrapfig}
\usepackage{xspace}
\usepackage{cuted}
\usepackage{graphicx}
\usepackage{amsmath}
\usepackage{amssymb}
\usepackage{float} % for [H] option if needed
\usepackage{ragged2e}
\usepackage{tabularx}

\usepackage[most]{tcolorbox}
\usepackage{xcolor}
\usepackage{times}
\usepackage{latexsym}
\usepackage{amsmath}
\usepackage{microtype}
\usepackage{multirow}
\usepackage{multirow}
\usepackage{graphicx}
\usepackage{graphicx}
\usepackage{enumitem}
\usepackage{xspace}
\PassOptionsToPackage{table}{xcolor}
\usepackage[table]{xcolor}
\usepackage{soul}
\usepackage{tabularx}
\usepackage{booktabs}
\usepackage{multirow}
\definecolor{lightpurple}{RGB}{244,238,252}
\definecolor{tableRow}{HTML}{F7F9FC}
\definecolor{tableHeaderBg}{HTML}{E8EEF7}
\usepackage{graphicx}
\usepackage{booktabs}
\usepackage{multirow}
\usepackage{array}
\usepackage{xcolor}
\usepackage{amssymb}

\title{\ours: Learning Trustworthy Memory Consolidation for LLM Agents with Long-Term Memory}
\author[1,2\dagger]{Tianyu Yang}
\author[1]{Sudipta Paul}
\author[1]{Vijay Srinivasan}
\author[1]{Vivek Kulkarni}
\author[1]{Srinivas Chappidi}

\affiliation[1]{AI Center-Mountain View, Samsung Electronics}
\affiliation[2]{University of Notre Dame}

\abstract{
Large language model (LLM) agents rely on long-term memory to support extended interactions and personalized assistance beyond finite context windows. Existing memory agents actively update external memory through \emph{generated write, revise}, and \emph{delete} operations, but these updates may omit important information, corrupt existing memory, or introduce unsupported hallucinated content. Once stored, such errors become persistent system-state failures that can affect future reasoning and generation. In this paper, we propose \textbf{\textsc{TrustMem}}, 
a framework designed to improve the trustworthiness of memory consolidation. TrustMem relies on a Memory Transition Verifier to evaluate the transition process of memory updates in terms of coverage, preservation, and faithfulness. It further constructs preference pairs among candidate updates under the same memory state, enabling preference-guided reinforcement learning to directly optimize memory updating behaviors.
Extensive experiments demonstrate that \ours improves both memory utility and reliability: it achieves state-of-the-art results across MemoryAgentBench, HaluMem, and the Mem-\(\alpha\) validation set, improves HaluMem memory extraction by 12.14 F1 points, and reduces transition-level omission, corruption, and hallucination by 40.1\%, 79.1\%, and 50.0\%, respectively, compared with the strongest baseline for each error type.

}

\begin{document}

\maketitle
\begingroup
\renewcommand{\thefootnote}{\fnsymbol{footnote}}
\footnotetext[2]{Work done during an internship at AI Center-Mountain View, Samsung Electronics.}
\endgroup

% \input{_sec_1_Introduction}
% \input{_sec_2_background}
% \input{_sec_3_Strategy}
% \input{_sec_4_Evaluation}
% \input{_sec_5_Challenges}
% \input{_sec_6_Conclusion}

% % \clearpage
% % \newpage
% % \bibliographystyle{assets/plainnat}
% \bibliographystyle{ieeenat_fullname}
% \bibliography{ijcai26}

\input{_sec_1_Introduction}

\input{_sec_2_background}
\input{_sec_3_Strategy}
\input{_sec_4_Evaluation}

\input{_sec_5_Challenges}

\bibliographystyle{ieeenat_fullname}
\bibliography{ijcai26}

\clearpage
\appendix
\input{_sec_6_Conclusion}
\end{document}

%% file: _sec_1_Introduction.tex
\section{Introduction}
% \vspace{-2mm}

\begin{figure*}[t]
\centering
\includegraphics[width=1\textwidth]{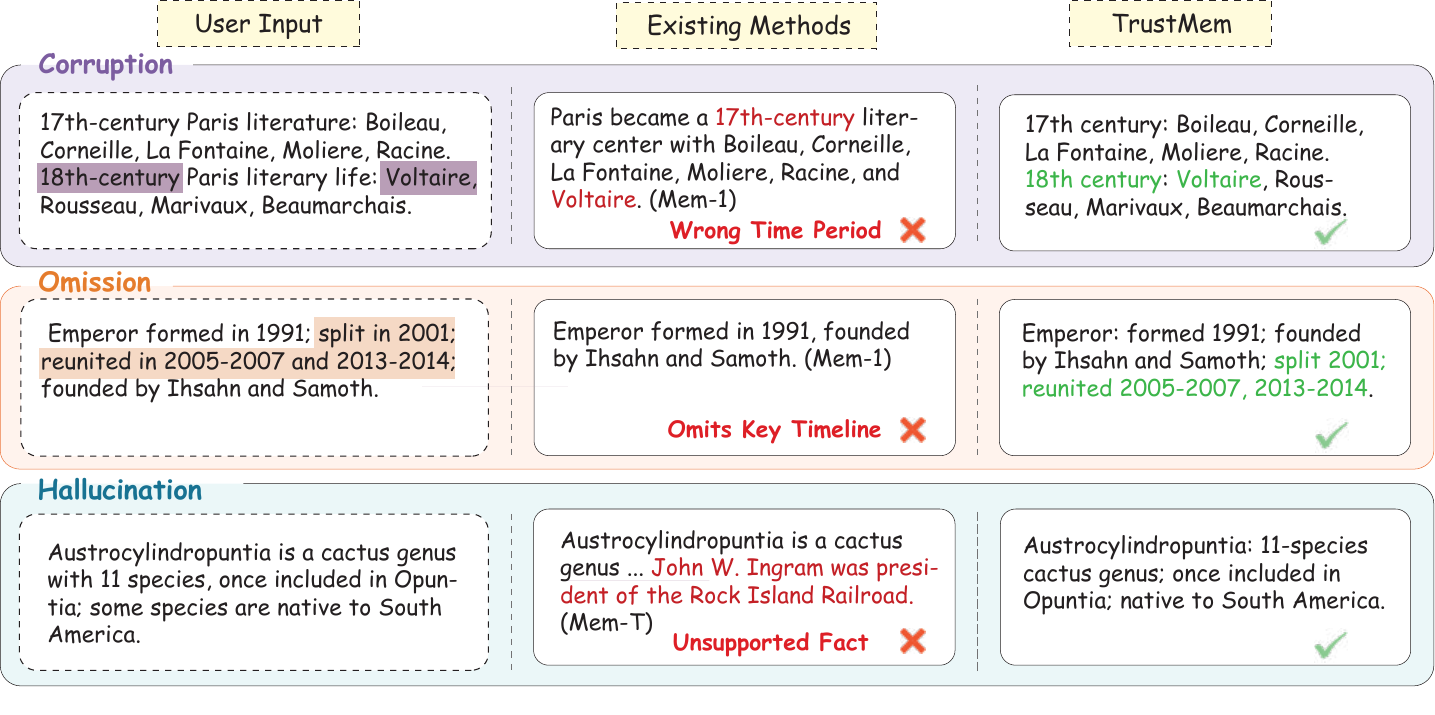}
\caption{
\textbf{Comparison of memory consolidation results produced by existing methods and \ours.}
% Existing memory agents may corrupt factual details, 
Compared with Mem1~\citep{zhou2025mem1} and Mem-T~\citep{yue2026mem}, \ours better preserves input evidence and avoids corruption, omission, and hallucination when converting user inputs into persistent memory.
}
%% \vspace{-20pt}
\label{fig:intro_case}
\end{figure*}

As Large Language Models (LLMs) evolve into intelligent agents capable of long-term interaction and personalized service, long-term memory has become a pivotal component~\cite{shinn2023reflexion, park2023generative}. By maintaining an external memory store, agents can transcend finite context windows to preserve user preferences, historical facts, and reasoning trajectories~\cite{zhong2024memorybank,lu2023memochat,packer2023memgpt}. Recent advancements~\cite{yu2025memagent,wang2025mem,yan2025memory,yue2026mem,zhou2025mem1} have moved beyond passive retrieval toward active memory management, where agents actively maintain and revise memory over time.
%treat memory as a dynamic, actionable state. 
Specifically, a memory agent typically processes streaming inputs in chunks: at each step, it retrieves relevant past contexts and executes structured memory-editing actions,  such as inserting new insights, updating existing records, or deleting obsolete information, to consolidate its persistent state. While this flexibility enhances controllability, it raises a critical concern: \textit{can we trust the autonomously generated memory updates?}

% \begin{figure}[!t]
% \centering
% \includegraphics[width=\columnwidth]{fig/2.2.png}
% % \vspace{-7mm}
% \caption{
% \textbf{Transition-level error rates of existing memory agents and \ours.}
% \ours achieves the lowest omission, corruption, and hallucination rates, reducing errors by 40.1\%, 79.1\%, and 50.0\% over the strongest baseline for each error type.
% }
% \label{fig:transition_level_errors}
% % \vspace{-8mm}
% \end{figure}

% \begin{figure*}[t]
% \centering
% \includegraphics[width=1\textwidth]{fig/4.7.pdf}
% \caption{
% \textbf{Comparison of memory consolidation results produced by existing methods and \ours.}
% % Existing memory agents may corrupt factual details, 
% Compared with Mem1~\citep{zhou2025mem1} and Mem-T~\citep{yue2026mem}, \ours better preserves input evidence and avoids corruption, omission, and hallucination when converting user inputs into persistent memory.
% }
% %% \vspace{-20pt}
% \label{fig:intro_case}
% \end{figure*}

% As shown in Figure~\ref{fig:intro_case}, existing memory agents can introduce persistent errors during memory consolidation even when given the same user inputs. 
As shown in Figure~\ref{fig:intro_case}, representative memory agents can introduce persistent errors during memory consolidation. These examples are drawn from outputs of Mem1~\citep{zhou2025mem1} and Mem-T~\citep{yue2026mem} under the same memory-consolidation setting used in our evaluation. 
For instance, the first row shows a \textbf{Corruption} error, where an existing method incorrectly places Voltaire in the 17th century despite the input associating him with the 18th century. The second row illustrates \textbf{Omission}, where key split and reunification timelines are dropped from memory. The third row shows \textbf{Hallucination}, where an unsupported statement about John W.~Ingram is introduced despite being grounded in neither the input nor the previous memory state. Since these errors can compound over time, the reliability of a memory system cannot be judged solely by terminal performance. Once a flawed entry is consolidated, it may be repeatedly retrieved and amplified in subsequent interactions, persistently contaminating the agent's reasoning chain~\cite{chen2025halumem}. Consequently, the core challenge of long-term memory systems has shifted from merely ``retaining'' information to \textit{building trustworthy memory}.
% This active autonomy introduces a fundamental challenge: unlike transient generation errors in standard dialogue, flawed memory updates permanently alter the agent's persistent state. During the update process, an agent may fail to capture critical new information (\textbf{Omission}), erroneously overwrite or distort existing knowledge (\textbf{Corruption}), or fabricate unsupported, hallucinated content (\textbf{Hallucination}). Because these errors compound over time, the reliability of a memory system cannot be judged solely by the terminal performance. Once a flawed entry is consolidated, it will be repeatedly retrieved and amplified in subsequent interactions, persistently contaminating the agent's reasoning chain. Consequently, the core challenge of long-term memory systems has shifted from merely "retaining" information to "building trustworthy memory."

% \vspace{-1mm}
Furthermore, existing supervisory signals struggle to address these issues due to a severe credit assignment gap~\cite{luo2025agent}. Most training paradigms~\cite{wang2025mem,yu2025memagent,yue2026mem,yan2025memory} rely on terminal scalar rewards—such as final task success or tool-calling validity—which act on the trajectory level rather than the transition level  for  memory updates. These global signals are too coarse to pinpoint exactly when or where a faulty memory update occurred within a long sequence of updates. An incorrect final answer could stem from an early omission, a destructive overwrite in mid-stream, or a simple reasoning flaw at the final step. Conversely, a correct answer may mask unsafe intermediate updates that did not trigger a failure in the current turn but remain latent risks for future tasks. Therefore, relying solely on terminal outcomes fails to provide the granular supervision necessary to explicitly correct the agent's specific memory-editing behaviors.

% To address these challenges, we formulate LLM memory management as a dynamic decision-making problem. However, optimizing this task-aware policy via standard Reinforcement Learning (RL) often exacerbates the credit assignment issue, as agents may learn spurious behaviors that satisfy terminal rewards but compromise memory integrity. To provide precise, step-level supervision, we introduce a \textbf{Memory Transition Verifier}. Breaking away from the reliance on terminal outcomes, the verifier acts as a dense reward model that scrutinizes the semantic safety of each isolated memory state change, explicitly penalizing omission, corruption, and hallucination before they can propagate through the long-term memory.

% Equipped with this dense supervisory signal, we propose \textbf{Transition-Ranked GRPO} to robustly optimize the agent's policy. This multi-granularity algorithm integrates global trajectory optimization with local preference ranking. On the global level, it utilizes Group Relative Policy Optimization (GRPO)~\cite{shao2024deepseekmath} to maximize downstream utility and action format compliance over the entire sequence. Simultaneously, on the local level, it leverages the verifier to rank candidate updates for the same chunk, explicitly steering the policy toward safe, verifier-preferred transitions. By bridging the credit assignment gap, our framework ensures the learned memory policy is both highly effective for long-term goals and fundamentally trustworthy at every action step.
To address this challenge, we shift the focus from evaluating only the final memory outcome to supervising how memory is constructed step by step. The agent processes the input chunk by chunk and executes structured memory-editing actions, such as writing new information, revising existing entries, or pruning outdated content. Each action sequence induces a local transition from the previous memory state to the updated one, enabling us to inspect whether each intermediate update is trustworthy before it becomes persistent.

Based on this formulation, we propose \textbf{\ours}, a framework for training memory agents to perform trustworthy memory consolidation at the transition level.
%a transition-level training framework for trustworthy memory consolidation. 
In particular, \ours introduces a \textbf{Memory Transition Verifier} to evaluate each memory  update along coverage, preservation, and faithfulness, and further proposes \textbf{Transition-Ranked GRPO}, built upon GRPO~\cite{shao2024deepseekmath}, to optimize the memory-update policy. By ranking candidate updates under the same memory state, \ours provides fine-grained supervision for memory-editing behaviors while maintaining the downstream utility of the final memory.

% We evaluate \ours across three settings: MemoryAgentBench~\cite{hu2025evaluating}, HaluMem~\cite{chen2025halumem}, and the Mem-\(\alpha\)~\cite{wang2025mem} validation set, which covers accurate retrieval, test-time learning, and long-range understanding. Extensive experiments demonstrate that TrustMem achieves state-of-the-art results across these settings, improving over the strongest baseline by 6.5 points on MemoryAgentBench and 12.14 F1 points in HaluMem memory extraction. Beyond final task performance, Figure~\ref{fig:transition_level_errors} 
% shows that TrustMem also achieves the lowest transition-level error rates across all three failure 
% types. Compared with the strongest baseline for each error type, TrustMem reduces omission, 
% corruption, and hallucination by 40.1\%, 79.1\%, and 50.0\%, respectively.These results demonstrate 
% that TrustMem improves not only memory utility, but also operation-level reliability and 
% transition-level safety.

\begin{figure}[!t]
\centering
\includegraphics[width=\columnwidth]{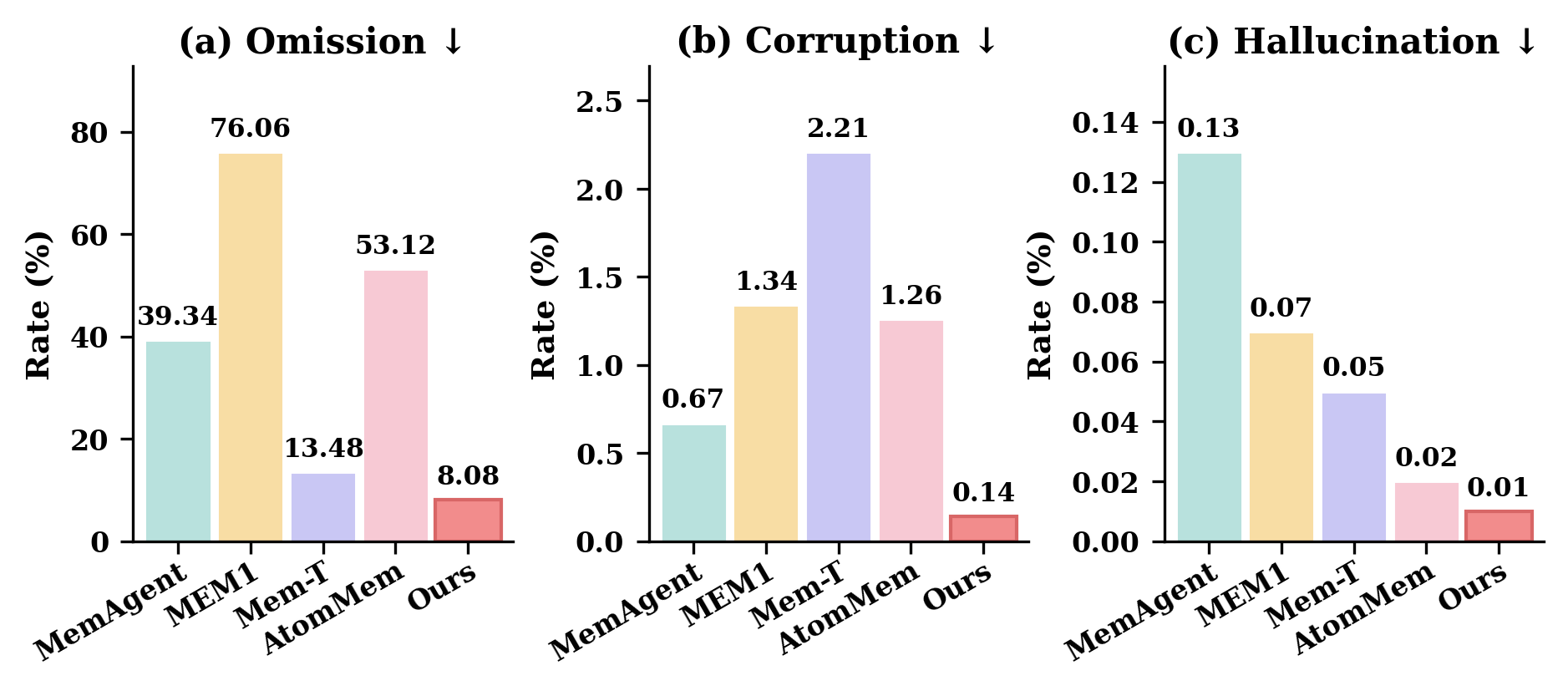}
% \vspace{-7mm}
\caption{
\textbf{Transition-level error rates of existing memory agents and \ours.}
\ours achieves the lowest omission, corruption, and hallucination rates, reducing errors by 40.1\%, 79.1\%, and 50.0\% over the strongest baseline for each error type.
}
\label{fig:transition_level_errors}
% % \vspace{-8mm}
\end{figure}

We evaluate \ours across three settings: MemoryAgentBench~\cite{hu2025evaluating}, 
HaluMem~\cite{chen2025halumem}, and the Mem-\(\alpha\)~\cite{wang2025mem} validation set, 
covering accurate retrieval, test-time learning, long-range understanding, and memory reliability. 
Extensive experiments demonstrate that \ours achieves state-of-the-art results across these 
settings, improving over the strongest baseline by 6.5 points on MemoryAgentBench and 12.14 F1 
points in HaluMem memory extraction. Beyond final task performance, Figure~\ref{fig:transition_level_errors} 
shows that \ours achieves the lowest transition-level error rates across omission, corruption, 
and hallucination. Compared with the strongest baseline for each error type, \ours reduces 
omission, corruption, and hallucination by 40.1\%, 79.1\%, and 50.0\%, respectively. These results 
demonstrate that \ours improves not only memory utility, but also operation-level reliability 
and transition-level safety.

%% file: _sec_2_background.tex
\section{Related Works}

% \vspace{-1mm}
\noindent
\textbf{Long-Term Memory Systems for LLM Agents.}
Long-term memory enables LLM agents to preserve information beyond finite context windows. 
Early systems store interactions, reflections, or user-specific notes in external memory, including Generative Agents~\citep{park2023generative}, Reflexion~\citep{shinn2023reflexion}, MemoryBank~\citep{zhong2024memorybank}, and MemoChat~\citep{lu2023memochat}. 
System-level frameworks such as MemGPT~\citep{packer2023memgpt}, MemoryOS~\citep{kang2025memory}, Mem0~\citep{chhikara2025mem0}, A-MEM~\citep{xu2025mem}, MIRIX~\citep{wang2025mirix}, Zep~\citep{rasmussen2025zep}, and LightMem~\citep{fang2025lightmem} improve memory management through external storage, extraction, organization, graph structure, retrieval, or efficiency. 
Another line explores parametric or latent memory, including MemoryLLM~\citep{wang2024memoryllm}, M+~\citep{wang2025m+}, Titans~\citep{behrouz2024titans}, and MemGen~\citep{zhang2025memgen}. 
These works primarily study how memory is stored, retrieved, compressed, or represented, whereas \ours focuses on whether each generated memory update is trustworthy before it gets stored.

\noindent
\textbf{RL-based Memory Agents.}
Recent methods formulate memory management as a learnable decision process. 
MemAgent~\citep{yu2025memagent} trains LLMs to process long inputs segment by segment and maintain compact memory; MEM1~\citep{zhou2025mem1} studies the interaction between memory construction and reasoning; Memory-R1~\citep{yan2025memory} learns explicit ADD, UPDATE, DELETE, and NOOP operations; and Mem-$\alpha$~\citep{wang2025mem} optimizes sequential memory construction over multi-component memories with RL. 
AgeMem~\citep{yu2026agentic}, ReMemR1~\citep{shi2025look}, and Mem-T~\citep{yue2026mem} further explore unified memory actions, revisitable memory, and denser credit assignment. 
While these methods demonstrate that RL can improve memory utility, they mainly optimize final task performance or trajectory-level rewards. 
\ours instead introduces transition-level verification and preference ranking to supervise the safety of each local memory update.

% % \vspace{-1mm}
\noindent
\textbf{Reliable Memory Agents.}
As memory becomes persistent and editable, reliability failures can accumulate across interactions. 
Benchmarks such as LoCoMo~\citep{maharana2024evaluating}, LongMemEval~\citep{wu2024longmemeval}, and MemoryAgentBench~\citep{hu2025evaluating} evaluate long-term memory through multi-session QA, temporal reasoning, knowledge updates, test-time learning, long-range understanding, or selective forgetting. 
HaluMem~\citep{chen2025halumem} further analyzes operation-level hallucination during memory extraction, updating, and memory-based QA. 
Related studies examine memory injection attacks~\citep{dong2025memory,tian2026injecmem}, memory admission control~\citep{zhang2026adaptive}, provenance-aware memory~\citep{zhu2026lossy}, and governance of evolving memory~\citep{lam2026governing}. 
Unlike these evaluation- or defense-oriented works, \ours directly optimizes the memory-update policy with transition-level verification, reducing omission, corruption, and hallucination before unsafe updates become persistent.

\begin{figure*}[t]
\centering
\includegraphics[width=1\textwidth]{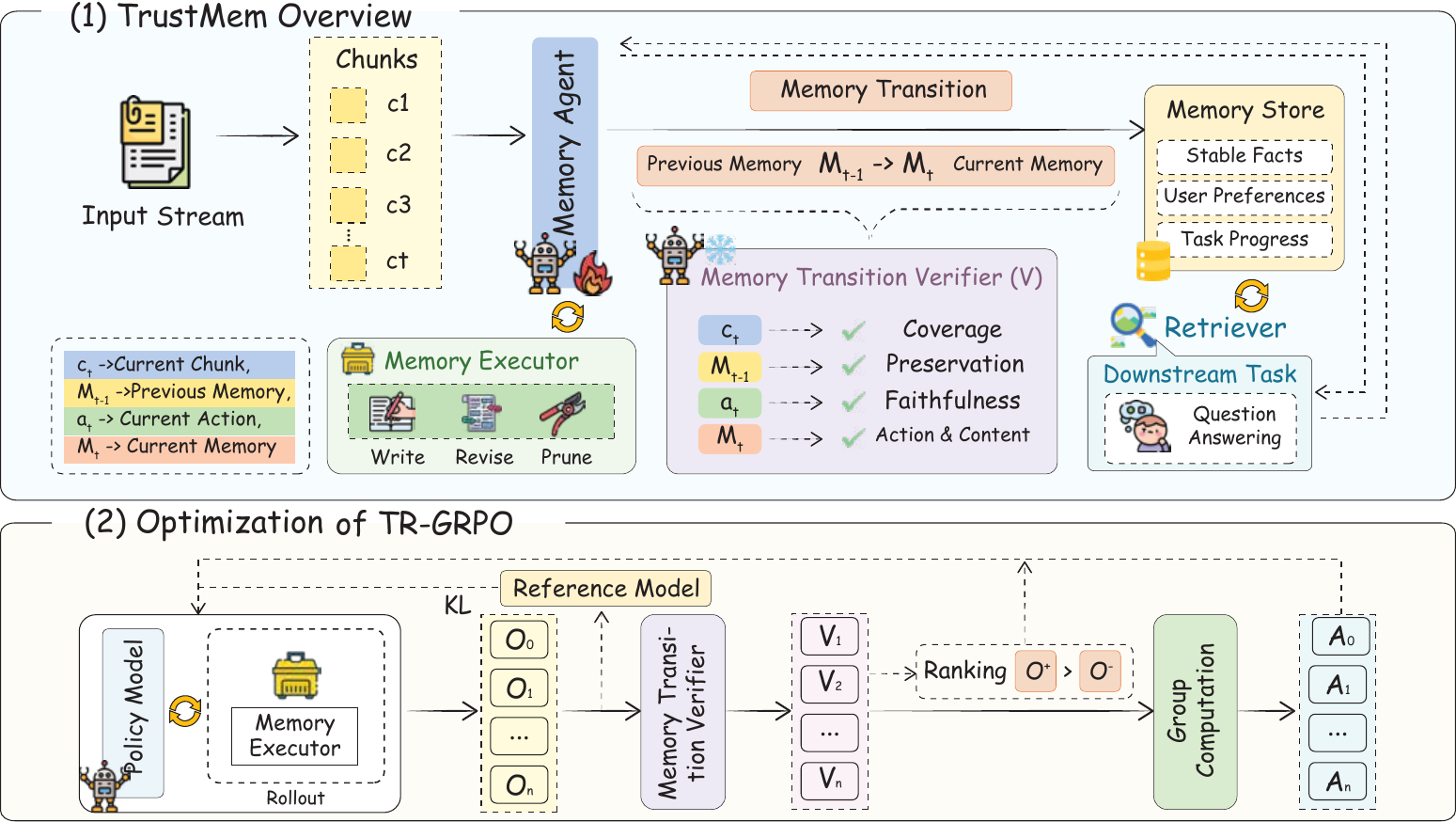}
% \vspace{-20pt}
\caption{
\textbf{Overview of \ours.}
\textbf{Top:} \ours performs continual memory consolidation over an input stream. 
At step \(t\), the memory agent observes the current chunk \(c_t\) and the accumulated memory state \(M_{t-1}\), generates memory-editing actions, and updates the memory state to \(M_t\) through the Memory Executor.
The resulting transition is evaluated by the Memory Transition Verifier for coverage, preservation, and faithfulness before the updated memory is used for retrieval and downstream tasks. 
\textbf{Bottom:} \ours optimizes the memory agent with Transition-Ranked GRPO, where verifier scores from multiple rollouts are used for both scalar reward computation and transition-level preference ranking.
% Top: \ours processes long-form inputs chunk by chunk. For each chunk, the memory agent generates structured memory-editing actions, and the Memory Executor applies them to write, revise, or prune entries in the external memory store. Each resulting memory transition is verified for coverage, preservation, and faithfulness before consolidation. Bottom: \ours optimizes the memory agent via Transition-Ranked GRPO, where verifier scores from multiple rollouts support both advantage computation and transition-level preference ranking.
}
% \vspace{-20pt}
\label{fig1}
\end{figure*}

%% file: _sec_3_Strategy.tex
\section{Method}
\label{sec:method}
\subsection{Overview of \ours Framework}
\label{sec:overview}

We formulate memory consolidation as a continual memory-update problem. 
At each step, the memory agent receives a new input chunk from a long-form interaction stream, such as a document segment, a dialogue turn, or a task demonstration segment. 
We denote the input stream as $C=\{c_1,c_2,\dots,c_T\}$, where $c_t$ is the newly observed chunk at step $t$. 
The chunks are processed in chronological order and consolidated into an evolving external memory.

Let $M_{t-1}$ denote the persistent memory state before processing chunk $c_t$. 
This memory state stores useful information accumulated from previous chunks $c_1,\dots,c_{t-1}$, rather than the content of only one chunk. 
At step $t$, the agent observes $(c_t, M_{t-1})$ and generates a sequence of structured memory-editing actions
$a_t \sim \pi_\theta(\cdot \mid c_t, M_{t-1})$.
Thus, the agent handles one chunk at each step, while the memory state represents global accumulated memory after all processed chunks so far.

% Given a long-form input sample $x$, we partition it into a sequence of chunks $C=\{c_1, c_2, \dots, c_T\}$. We aim to train a parameterized memory agent $A_\theta$ (with policy $\pi_\theta$) that sequentially processes these chunks to maintain an external memory state $M_t$. Unlike temporary model context, this persistent memory store consists of editable entries---such as stable facts, user preferences, and task progress---that can be dynamically managed across long-horizon interactions.

% At step $t$, the agent observes the current chunk $c_t$ and the previous memory state $M_{t-1}$, generating a sequence of structured memory-editing actions $a_t \sim \pi_\theta(\cdot \mid c_t, M_{t-1})$. To make updates executable and auditable, we define a unified action space $\mathcal{A}=\{\textsc{Write}, \textsc{Revise}, \textsc{Prune}\}$. Specifically, \textsc{Write} inserts new memory items, \textsc{Revise} modifies or merges existing ones, and \textsc{Prune} removes outdated content. The action sequence can contain multiple edits, or remain empty if the current chunk requires no update. A deterministic memory executor $\mathcal{E}$ then applies $a_t$ to the memory store, yielding the updated state $M_t=\mathcal{E}(M_{t-1},a_t)$. This incremental construction defines an explicit memory transition $z_t=(c_t, M_{t-1}, a_t, M_t)$.

To make memory updates executable and auditable, we define a unified action space
$\mathcal{A}=\{\textsc{Write}, \textsc{Revise}, \textsc{Prune}\}$.
Specifically, \textsc{Write} inserts new memory items when the current chunk contains useful information not yet stored in memory; 
\textsc{Revise} updates or merges existing memory entries when the current chunk provides corrections, refinements, or complementary evidence; 
and \textsc{Prune} removes memory entries that become invalid, redundant, or no longer useful for downstream tasks based on the current evidence and the existing memory state.
The agent decides whether an update is needed by generating one or more of these actions conditioned on $(c_t,M_{t-1})$. 
If the current chunk does not introduce new useful information, correct previous memory, or make any existing entry invalid or redundant, the agent may output no action.

A deterministic memory executor $\mathcal{E}$ then applies $a_t$ to the memory store, yielding the updated state
$M_t=\mathcal{E}(M_{t-1},a_t)$.
This incremental construction defines a local memory transition
$z_t=(c_t,M_{t-1},a_t,M_t)$.

Since each update alters the agent's persistent memory state, we introduce a Memory Transition Verifier $V$ to evaluate the trustworthiness of each local transition, yielding a score $v_t=V(z_t) \in [0,1]$. 
As detailed in Section~\ref{sec:verifier}, the verifier checks if the update preserves important information from the current chunk, maintains valid prior memory, and avoids unsupported hallucinated content.

After all $T$ chunks are processed, the terminal memory $M_T$ is used for downstream tasks. 
For a query $q$, a fixed retriever fetches relevant records $S_q=\mathcal{R}(M_T,q)$, and an answer model generates a prediction $\hat{y}=g_{\text{ans}}(q,S_q)$, which is compared against the reference answer $y$ to compute the task reward. 
Our overall objective is to optimize the policy $\pi_\theta$ so that the agent constructs compact and useful terminal memory while ensuring that each intermediate memory transition is trustworthy.

% \vspace{-1mm}
\subsection{Memory Transition Verifier}
\label{sec:verifier}
% \vspace{-1mm}

Because memory construction is cumulative, relying solely on terminal downstream accuracy suffers from a severe credit assignment gap. 
An incorrect final answer may stem from an early omission, a later destructive edit, or a failure of the answer model. 
Conversely, a correct final answer does not guarantee that all intermediate memory updates are safe, since silent errors may remain in the persistent memory state and affect future tasks.

To provide fine-grained supervision, we introduce a \textbf{Memory Transition Verifier} $V$ that evaluates each local memory transition defined in Section~\ref{sec:overview}. 
Rather than judging the final answer, the verifier acts as a state-transition evaluator and checks whether the update from $M_{t-1}$ to $M_t$ is trustworthy after processing the current chunk.

For each transition $z_t=(c_t,M_{t-1},a_t,M_t)$, the verifier is instantiated as a frozen LLM and receives a compact view of the transition, including the current chunk, the generated memory-editing actions, and the memory entries touched or retrieved by these actions. 
This avoids feeding the full memory bank while still providing the evidence needed to judge the local update.
The simplified verifier prompt is shown in Figure~\ref{fig:transition-verification-prompt} in the Appendix.

Specifically, the verifier evaluates each transition along three diagnostic dimensions:
\textbf{coverage}, which checks whether important information from the current chunk is preserved;
\textbf{preservation}, which checks whether valid prior memory is maintained without unjustified deletion or distortion;
and \textbf{faithfulness}, which checks whether newly added or modified memory content is supported by the current chunk or the previous memory state.
These judgments are aggregated into a scalar trustworthiness score
\(v_t = V(z_t) \in [0,1]\), which is used as the transition-level verification reward.

By directly evaluating intermediate memory transitions, the verifier provides local supervision for memory-editing behavior and helps prevent omission, corruption, and hallucination before they become persistent memory errors.

\subsection{Multi-granularity Reward Design}
\label{sec:reward}

Training reliable memory agents requires supervision at different granularities. 
Sample-level rewards evaluate the utility and compactness of the final memory after the full input stream has been processed, while transition-level rewards evaluate whether each local memory update is executable, meaningful, and trustworthy. 
We therefore design a multi-granularity reward that combines sample-level rewards with transition-level rewards.

\noindent
\textbf{Sample-Level Rewards.}
These rewards evaluate the terminal memory state $M_T^{(i)}$ after all chunks in sample $i$ are processed. 

\noindent
\textit{\underline{Task Reward}} ($R_{\text{task}}$) measures downstream utility: given questions $Q^{(i)}$, a fixed retriever $\mathcal{R}$ and answer model $g_{\text{ans}}$ generate predictions $\hat{y}$, and the reward is the average metric, such as F1 or LLM-as-a-judge score, against reference answers $y$. 

\noindent
\textit{\underline{Efficiency Reward}} ($R_{\text{eff}}$) penalizes memory bloat to prevent excessive copying, computed as $1 - (|M_T^{(i)}| / \sum_t |c_t^{(i)}|)$, encouraging compact memory.

\noindent
\textbf{Transition-Level Rewards.}
These rewards evaluate each local memory transition $M_{t-1}\rightarrow M_t$. 

\noindent
\textit{\underline{Action Execution}} ($R_{\text{act}}$) checks whether the generated structured actions are parseable and executable by the memory executor $\mathcal{E}$.

\noindent
\textit{\underline{Content Type}} ($R_{\text{type}}$) uses a lightweight analyzer $J_{\text{type}}$ to ensure that the modified content is semantically specific and non-empty. 

\noindent
\textit{\underline{Verifier Reward}} ($R_{\text{ver}}$) uses the Memory Transition Verifier $V$ from Section~\ref{sec:verifier} to evaluate the semantic safety of the transition, defined as $R_{\text{ver},t}^{(i)} = V(z_{i,t})$.

Aggregating the above, we define the total reward for the $t$-th memory update of sample $i$ as
% \vspace{-1mm}
\begin{equation}
  \label{eq:total_reward}
  \begin{aligned}
  R_{i,t} & =
  \; R_{\text{task}}^{(i)}
  + \lambda_{\text{eff}} R_{\text{eff}}^{(i)}
  &+ \lambda_{\text{act}} R_{\text{act},t}^{(i)}
  + \lambda_{\text{type}} R_{\text{type},t}^{(i)}
  + \lambda_{\text{ver}} R_{\text{ver},t}^{(i)} .
  \end{aligned}
\end{equation}
% \vspace{-1mm}
This reward combines final memory utility, memory compactness, action executability, content specificity, and transition-level trustworthiness. 
The sample-level terms provide global supervision over the final constructed memory, while the transition-level terms provide local supervision for each memory update.

% \vspace{-1mm}
\subsection{Transition-Ranked GRPO}
\label{sec:tr-grpo}

Building on the multi-granularity reward, we propose \textbf{Transition-Ranked GRPO}, which combines scalar reward optimization with verifier-guided transition ranking. 
While the verifier score can be used as a transition-level reward in Eq.~\eqref{eq:total_reward}, its absolute magnitude may not always be well calibrated. 
However, in our setting, verifier scores are used more robustly as relative preferences among candidate updates generated under the same state than as fully calibrated absolute values.
Therefore, we further use verifier scores to construct preference pairs among candidate transitions and optimize the policy with an additional ranking objective.

\paragraph{Global Trajectory Optimization via GRPO.}
We adopt Group Relative Policy Optimization (GRPO) \cite{shao2024deepseekmath} to optimize the memory policy. 
Specifically, for each training sample $i$, we sample a group of $G$ rollouts from the old policy $\pi_{\theta_{\text{old}}}$. 
At each chunk step $t$ within rollout $g$, we compute the scalar reward $R_{i,t}^{(g)}$ from Eq.~\eqref{eq:total_reward}. 
The group-normalized advantage is computed as 
$\hat{A}_{i,t}^{(g)}=(R_{i,t}^{(g)}-\mu_{i,t})/(\sigma_{i,t}+\epsilon_A)$, 
where $\mu_{i,t}$ and $\sigma_{i,t}$ are the mean and standard deviation of rewards across the $G$ rollouts for the same chunk step. 
Letting $r_{i,t,l}^{(g)}(\theta)$ denote the importance ratio for the $l$-th generated token, the GRPO objective over the memory construction trajectory is formulated as:
% \vspace{-2mm}
\begin{equation}
  \label{eq:grpo_objective}
  \begin{aligned}
    \mathcal{J}_{\text{GRPO}}(\theta)
    &=
    \mathbb{E}_{i,t,g,l} \Big[
    \min \big(
    r_{i,t,l}^{(g)}(\theta)\hat{A}_{i,t}^{(g)}, 
    &\quad
    \text{clip}\big(r_{i,t,l}^{(g)}(\theta), 1-\epsilon, 1+\epsilon\big)
    \hat{A}_{i,t}^{(g)}
    \big)
    \Big].
  \end{aligned}
\end{equation}
% \vspace{-6mm}
% \begin{equation}
%   \label{eq:grpo_objective}
%   \begin{aligned}
%     \mathcal{J}_{\text{GRPO}}&(\theta) = \mathbb{E} \Bigg[ \frac{1}{N} \sum_{i,t,g,l} \min \Big( r_{i,t,l}^{(g)}(\theta)\hat{A}_{i,t}^{(g)}, \\
%     &\quad \text{clip}\big(r_{i,t,l}^{(g)}(\theta), 1-\epsilon, 1+\epsilon\big) \hat{A}_{i,t}^{(g)} \Big) \Bigg].
%   \end{aligned}
%   % \vspace{-5mm}
% \end{equation}

\paragraph{Verifier-Guided Pair Construction.}
For each training sample $i$, the $G$ rollouts produce multiple candidate memory transitions at each chunk step $t$, denoted as $\mathcal{Z}_{i,t}=\{z_{i,t}^{(1)},\dots,z_{i,t}^{(G)}\}$. These candidate transitions are generated from the same current chunk and previous memory state, but may contain different memory-editing actions. 
Because the verifier $V$ evaluates each transition independently, it induces a ranking over $\mathcal{Z}_{i,t}$. 
We select the transitions with the highest and lowest verifier scores as the preferred transition $z^+$ and rejected transition $z^-$, respectively, provided their score gap exceeds a margin $\delta$. 
These preference pairs capture local differences in transition quality and provide additional supervision beyond the scalar reward.
% \vspace{-2mm}
\paragraph{Local Preference Ranking via Verifier.}
Given the preference pairs $(z^+, z^-)$, we optimize a DPO-style ranking loss to inject step-level semantic supervision. 
Because $z^+$ and $z^-$ are generated for the same chunk and previous memory state, they share the same prompt $p_{i,t}$ and differ only in the generated memory-editing actions. 
Let $o^+$ and $o^-$ denote the preferred and rejected action sequences. 
% We define the length-normalized log-probability margin as
% \[
% \Delta_\theta = \ell_\theta(o^+\mid p_{i,t}) - \ell_\theta(o^-\mid p_{i,t}),
% \]
% \input{tables/tab_3}
We define the length-normalized log-probability margin as $\Delta_\theta = \ell_\theta(o^+\mid p_{i,t}) - \ell_\theta(o^-\mid p_{i,t})$, 
where $\ell_\theta(o\mid p)$ denotes the length-normalized log probability of a generated response $o$ given prompt $p$. 
The ranking loss is:
\begin{equation}
  \label{eq:ranking_loss}
  \mathcal{L}_{\text{rank}} = -\log \sigma \left( \beta_{\text{rank}} \left[ \Delta_\theta - \Delta_{\text{ref}} \right] \right),
  % % \vspace{-5pt}
\end{equation}
where $\Delta_{\text{ref}}=\ell_{\mathrm{ref}}(o^+\mid p_{i,t})-\ell_{\mathrm{ref}}(o^-\mid p_{i,t})$ is the corresponding margin under the reference policy.

\paragraph{Joint Objective.}
We use a frozen reference policy $\pi_{\mathrm{ref}}$ to regularize policy updates with a KL penalty. 
This KL term is applied as an actor-side auxiliary loss, rather than being added to the scalar reward used for advantage estimation. 
Here, $\pi_{\theta_{\mathrm{old}}}$ denotes the rollout policy used in the GRPO importance ratio, while $\pi_{\mathrm{ref}}$ denotes the frozen reference policy used for KL regularization and the ranking margin.

The final training objective combines scalar reward optimization, KL regularization, and transition-level ranking:
% \vspace{-1mm}
\begin{equation}
  \label{eq:joint_objective}
  \mathcal{L}
  =
  -\mathcal{J}_{\text{GRPO}}
  + \beta_{\text{KL}}\mathcal{L}_{\text{KL}}
  + \lambda_{\text{rank}}\mathcal{L}_{\text{rank}}.
\end{equation}
% \vspace{-1mm}
Here, $\mathcal{L}_{\text{KL}}$ denotes the token-level KL regularization loss, while $\beta_{\text{KL}}$ and $\lambda_{\text{rank}}$ control the weights of KL regularization and transition-level ranking, respectively.

\input{tables/tab_1}

\input{tables/tab_2}

%% file: tables/tab_1.tex
\begin{table*}[t]
    \centering
    \small
    \resizebox{\textwidth}{!}{
    \begin{tabular}{l|ccc|ccc|c|c}
    \toprule
    \multirow{2}{*}{\textbf{Method}}
         & \multicolumn{3}{c|}{\textbf{AR}} 
         & \multicolumn{3}{c|}{\textbf{TTL}} 
         & \textbf{LRU} 
         & \multirow{2}{*}{\textbf{Avg.}} \\
    \cmidrule(lr){2-4}
    \cmidrule(lr){5-7}
    \cmidrule(lr){8-8}
         & SQuAD & HotpotQA & PerLTQA 
         & TREC-C & NLU & PubMed 
         & BookSum & \\
    \midrule

        Long-Context
            & 74.2 & \textbf{85.2} & 60.5 
            & 62.3 & \underline{70.8} & 53.3 
            & 5.2 & 58.8 \\

        RAG-Top2
            & 76.2 & \underline{84.9} & 62.3 
            & 61.2 & 50.8 & 57.0 
            & 4.2 & 56.7 \\

        MemAgent~\citep{yu2025memagent}
            & 52.6 & 66.0 & 41.5 
            & 53.5 & 57.4 & \underline{61.6} 
            & 8.6 & 48.7 \\

        MEM1~\citep{zhou2025mem1}
            & 12.0 & 16.4 & 16.0 
            & 47.5 & 33.2 & 22.6 
            & 6.6 & 22.0 \\

        Mem-$\alpha$~\citep{wang2025mem}
            & \underline{78.6} & 83.2 & \textbf{65.9}
            & \underline{66.6} & 65.8 & 54.5
            & \underline{18.7} & \underline{61.9} \\

        \rowcolor{lightpurple}
        \ours
            & \textbf{80.6} & \underline{84.9} & \underline{63.6} 
            & \textbf{72.4} & \textbf{73.8} & \textbf{64.0} 
            & \textbf{22.6} & \textbf{66.3} \\

    \bottomrule
    \end{tabular}
    }
    \vspace{-2mm}
    \caption{Experiment results on the Mem-\(\alpha\) validation set. The top-2 results are highlighted.
    }
    \vspace{-2mm}
    \label{tab:main_result_mem-alpha}
\end{table*}

%% file: tables/tab_2.tex
\begin{table*}[t]
    \centering
    \small
    \setlength{\tabcolsep}{3pt}
    \resizebox{\textwidth}{!}{
    \begin{tabular}{l|ccc|ccccc|c|c}
    \toprule
    \multirow{2}{*}{\textbf{Method}}
         & \multicolumn{3}{c|}{\textbf{AR}} 
         & \multicolumn{5}{c|}{\textbf{TTL}} 
         & \textbf{LRU} 
         & \multirow{2}{*}{\textbf{Avg.}} \\
    \cmidrule(lr){2-4}
    \cmidrule(lr){5-9}
    \cmidrule(lr){10-10}
         & Single-Doc & Multi-Doc & LME(S) 
         & TREC-C & NLU & TREC-F & Clinic & Banking77 
         & InfBench & \\
    \midrule

        Long-Context 
            & 28.0 & 27.0 & 29.2 
            & 64.0 & 74.0 & 34.0 & \underline{86.0} & 77.0 
            & 12.5 & 46.1 \\

        RAG-Top2 
            & 69.0 & 45.0 & \textbf{58.1} 
            & 69.0 & 65.0 & 21.0 & 70.0 & 75.0 
            & 6.5 & 50.2 \\

        MemAgent~\citep{yu2025memagent}
            & 51.0 & 52.0 & 23.0 
            & 62.0 & 78.0 & 30.0 & \textbf{88.0} & \underline{83.0} 
            & 10.9 & 53.1 \\

        MEM1~\citep{zhou2025mem1}
            & 23.0 & 33.0 & 13.0 
            & 57.0 & \textbf{80.0} & 28.0 & 48.0 & 76.0 
            & 7.0 & 40.6 \\

        Mem-T~\citep{yue2026mem}
            & 56.0 & 60.0 & 43.3
            & \underline{75.0} & 48.0 & 27.0 & 43.0 & 79.0
            & 14.1 & 49.5 \\

        AtomMem~\citep{huo2026atommem}
            & 37.0 & 33.0 & 8.7
            & 11.0 & 1.0 & 9.0 & 0.0 & 0.0
            & \textbf{18.9} & 13.2 \\

        Mem-$\alpha$~\citep{wang2025mem}
            & \underline{74.0} & \textbf{68.0} & 52.0 
            & 71.0 & 71.0 & \textbf{41.0} & 73.0 & 70.0 
            & 12.9 & \underline{59.2} \\

        \rowcolor{lightpurple}
        \ours
            & \textbf{83.0} & \underline{64.0} & \underline{55.0} 
            & \textbf{85.0} & \underline{79.2} & \textbf{41.0} & 84.0 & \textbf{84.0} 
            & \underline{15.9} & \textbf{65.7} \\

    \bottomrule
    \end{tabular}}
    \vspace{-2mm}
    \caption{Experiment results on MemoryAgentBench. The top-2 results are highlighted.}
    \vspace{-5mm}
    \label{tab:main_result_memoryagentbench}
\end{table*}

%% file: _sec_4_Evaluation.tex
% % % % % % \vspace{-1mm}
\section{Experiment}
\label{sec:experiment}
% % % % % % \vspace{-1mm}

% \subsection{Experiment Settings}
% \label{sec:exp_settings}

% We provide experimental settings, including dataset statistics, baseline descriptions, implementation details, and evaluation protocols, in Appendix~\ref{setup}.

\subsection{Experiment Settings}
\label{setup}
\noindent\textbf{Datasets.} 
We evaluate \ours on MemoryAgentBench~\cite{hu2025evaluating}, HaluMem~\cite{chen2025halumem}, and the Mem-\(\alpha\) validation set. 
Following Mem-\(\alpha\)~\citep{wang2025mem}, we use its training pool, which aggregates eight sources: SQuAD~\citep{rajpurkar2016squad}, HotpotQA~\citep{yang2018hotpotqa}, PerLTQA~\citep{du2024perltqa}, LongMemEval-Train~\citep{wu2024longmemeval}, NLU~\citep{larson2019evaluation}, TREC-C~\citep{li2006learning}, PubMed-RCT~\citep{dernoncourt2017pubmed}, and BookSum~\citep{kryscinski2022booksum}. We train \ours on a balanced subset of 562 instances sampled from the 4,139-instance Mem-\(\alpha\) training pool, and evaluate validation performance and ablations on 463 held-out Mem-\(\alpha\) validation instances. We use all datasets, models, and systems only for research purposes and follow their publicly released access conditions and licenses where available.

\input{tables/tab_3}
% \noindent\textbf{Datasets.} 
% We evaluate \ours on two long-term memory benchmarks: MemoryAgentBench~\cite{hu2025evaluating} and HaluMem~\cite{chen2025halumem}, and additionally report results on the validation set. Following prior work~\cite{wang2025mem}, our training data aggregates eight sources ((SQuAD~\citep{rajpurkar2016squad}, HotpotQA~\citep{yang2018hotpotqa}, PerLTQA~\citep{du2024perltqa}, LongMemEval-Train~\citep{wu2024longmemeval}, NLU~\citep{larson2019evaluation}, TREC-C~\citep{li2006learning}, PubMed-RCT~\citep{dernoncourt2017pubmed}, and BookSum~\citep{kryscinski2022booksum})), covering accurate retrieval, test-time learning, and long-range understanding. Out of 4,139 total training instances, we use a balanced sample of 562 instances to mitigate data imbalance and RL training costs, while retaining 463 instances for validation.

\noindent\textbf{MemoryAgentBench.}
We evaluate \ours on representative tasks from MemoryAgentBench~\cite{hu2025evaluating} across three memory capabilities. For Accurate Retrieval (AR), we use Single-Doc, Multi-Doc, and LME(S), which evaluate whether the agent can retrieve useful information from historical contexts for downstream queries. For Test-Time Learning (TTL), we use five multi-class classification datasets: TREC-C~\citep{li2006learning}, NLU~\citep{larson2019evaluation}, TREC-F~\citep{li2006learning}, Clinic~\citep{larson2019evaluation}, and Banking77~\citep{casanueva2020efficient}, which test whether the agent can acquire new classification rules from examples during interaction. For Long-Range Understanding (LRU), we use InfBench-Sum~\citep{zhang2024bench}, which evaluates the agent's ability to integrate information across long contexts for summarization. Each task is evaluated with its original task-specific scoring protocol, and the final results are reported as task-level performance scores. 

\noindent\textbf{HaluMem Bench.} 
HaluMem~\cite{chen2025halumem} is an operation-level benchmark focusing on memory safety, including hallucinations, omissions, and incorrect updates, during memory extraction, updating, and question answering. We use HaluMem-Medium, which contains 30,073 dialogue turns from 20 users ($\sim$160K tokens/user), encompassing 14,948 memory points and 3,467 QA pairs. While MemoryAgentBench evaluates downstream memory utility and out-of-distribution generalization, HaluMem evaluates memory update safety and operation-level reliability through stage-wise extraction, updating, and QA metrics.

\noindent
\textbf{Baselines.}
We compare \ours  with a broad set of baselines covering three memory-processing paradigms, as summarized in Tables~\ref{tab:baseline_model_details}.
(1) \textit{Standard paradigms}: \textbf{Long-Context} directly places the interaction history into the context window, while \textbf{RAG-Top2} retrieves the top-2 relevant chunks before answer generation. These baselines test whether long-context encoding or simple retrieval is sufficient for long-term memory tasks.
(2) \textit{RL-based and agentic memory methods}: We compare with recent methods that train or design agents to construct, update, and use long-term memory, including \textbf{Mem-$\alpha$}~\cite{wang2025mem}, \textbf{MemAgent}~\cite{yu2025memagent}, \textbf{MEM1}~\cite{zhou2025mem1}, \textbf{Mem-T}~\cite{yue2026mem}, and \textbf{AtomMem}~\cite{huo2026atommem}. These methods typically process long inputs incrementally and maintain compact or structured memory states for downstream tasks.
(3) \textit{System-level memory frameworks}: We further compare with practical memory management systems, including \textbf{Mem0}~\citep{chhikara2025mem0}, \textbf{Memobase}~\citep{memobase2025}, \textbf{Supermemory}~\citep{supermemory2025}, and \textbf{Zep}~\citep{rasmussen2025zep}, which provide memory extraction, updating, retrieval, and memory-based question answering capabilities.
Since different benchmarks emphasize different aspects of memory behavior and not all baselines are applicable to every setting, we report the available and comparable results for each benchmark in the corresponding tables.

\noindent
\textbf{Implementation Details.}
\ours is initialized from Qwen3-4B~\cite{yang2025qwen3} and trained only on the Mem-\(\alpha\) training pool with Transition-Ranked GRPO for approximately two days on 32 NVIDIA H100 GPUs. 
No examples from MemoryAgentBench or HaluMem are used for training; all results are reported on held-out validation instances or external benchmarks. We report results from the step-85 checkpoint. Following the reward definition in Section~\ref{sec:method}, the training reward is defined as \(R = R_{\text{task}} + R_{\text{act}} + 0.1 R_{\text{ver}} + 0.05 R_{\text{eff}} + 0.1 R_{\text{type}}\), where \(R_{\text{task}}\) measures downstream task performance, \(R_{\text{act}}\) evaluates action executability, \(R_{\text{ver}}\) measures transition-level memory trustworthiness, \(R_{\text{eff}}\) encourages compact memory construction, and \(R_{\text{type}}\) evaluates content specificity. We use 8 GRPO rollouts per instance, a learning rate of \(1\times10^{-6}\), KL coefficient 0.001, maximum prompt length 4096, maximum response length 2048. For the preference-ranking objective, we use top-1/bottom-1 pair construction with score margin 0.15, at most 256 preference pairs per batch, pair micro-batch size of 1 per GPU, preference loss weight \(\lambda_{\text{rank}}=0.05\), and \(\beta_{\text{rank}}=0.1\). Thinking mode is disabled during both training and evaluation.

\noindent
\textbf{Evaluation Metrics.}
For the MemoryAgentBench and validation set , we follow the original evaluation protocols of the corresponding benchmarks. Since different data sources use different scoring functions, we evaluate each instance with its dataset-specific metric and report the resulting task-level performance score for each dataset. The Avg. column denotes the macro-average over the reported task scores. For HaluMem, we report operation-level metrics over three stages: memory extraction, memory updating, and memory-based question answering. For memory extraction, we report recall (R), weighted recall (Weighted R), accuracy (Acc.), and F1. For memory updating and question answering, we report correctness (C), hallucination rate (H), and omission rate (O). Higher values indicate better performance for R, Weighted R, Acc., F1, and C, while lower values are better for H and O.

\subsection{Experiments on Memory Utility}
\label{sec:exp_memory_utility}

Table~\ref{tab:main_result_mem-alpha} reports results on the Mem-\(\alpha\) validation set. TrustMem achieves the best average score of 66.3, outperforming the strongest baseline Mem-\(\alpha\) by 4.4 points. The improvement is especially clear on Test-Time Learning, where TrustMem obtains the best scores on TREC-C (72.4), NLU (73.8), and PubMed (64.0). TrustMem also improves BookSum from 18.7 to 22.6, showing better long-range memory utility. These gains indicate that transition-level verification helps the agent preserve useful task-specific and long-range information during memory construction.

Table~\ref{tab:main_result_memoryagentbench} evaluates out-of-distribution generalization on MemoryAgentBench. TrustMem again achieves the best average score of 65.7, improving over Mem-\(\alpha\) by 6.5 points and outperforming Long-Context and RAG-Top2 by 19.6 and 15.5 points, respectively. TrustMem performs strongly on both retrieval and test-time learning tasks, achieving the best score on Single-Doc (83.0), TREC-C (85.0), and Banking77 (84.0). These results show that TrustMem learns a more robust memory update policy than simply retaining long context or retrieving top-ranked chunks. Overall, these results demonstrate that reliable local memory transitions improve both validation performance and out-of-distribution memory utility, with detailed analysis provided in Appendix~\ref{sec:exp_memory_utility_appendix}.

\subsection{Experiments on HaluMem}
\label{sec:exp_halumem}

Table~\ref{tab:halumem_medium_full_results} evaluates operation-level memory reliability on HaluMem-Medium, covering memory extraction, memory updating, and memory-based question answering. 
TrustMem achieves the best memory extraction recall (56.73), weighted recall (67.45), and F1 (69.45), improving the strongest prior F1 score by 12.14 points. 
Although AtomMem obtains the highest extraction accuracy, its very low recall and F1 suggest an overly conservative extraction behavior that misses most useful memories. 
In contrast, TrustMem achieves a stronger coverage-correctness trade-off, resulting in the best overall extraction quality. For memory updating and memory-based question answering, TrustMem obtains the lowest omission rates while maintaining competitive correctness and low hallucination. 
These results show that transition-level verification and preference-guided optimization improve intermediate memory reliability beyond final task performance. 
A detailed stage-wise analysis is provided in Appendix~\ref{sec:exp_halumem_appendix}.

\subsection{Ablation Study}
\label{sec:ablation}

As shown in Table~\ref{tab:ablation}, we perform ablation studies on the Mem-\(\alpha\) validation set to quantitatively evaluate the impact of Action Execution Reward (\(R_{\text{act}}\)), Efficiency Reward (\(R_{\text{eff}}\)), Transition Verifier Reward (\(R_{\text{ver}}\)), and the Transition-Ranked GRPO optimization strategy.

\input{tables/tab_4}

%We analyze three key aspects of TrustMem: reward components, the Memory Transition Verifier, and the optimization strategy.

\noindent
\textbf{Effectiveness of Reward Components.}
Removing \(R_{\text{act}}\) causes the largest drop, reducing the overall score from 66.3 to 54.0. This shows that executable memory-editing actions are crucial for reliable memory construction. Removing \(R_{\text{eff}}\) also decreases the overall score to 61.8, indicating that compact memory states help reduce redundancy and improve downstream memory utility.

\noindent
\textbf{Effectiveness of the Memory Transition Verifier.}
Removing \(R_{\text{ver}}\) reduces the overall score from 66.3 to 58.7, with a clear drop on TTL from 70.1 to 58.3. This confirms that the Memory Transition Verifier provides important transition-level supervision. By checking local memory updates for omission, corruption, and hallucination, the verifier helps prevent unsafe updates from entering the persistent memory state.

\noindent
\textbf{Effectiveness of Transition-Ranked GRPO.}
Compared with PRM-style GRPO, TrustMem with Transition-Ranked GRPO improves the overall score from 60.3 to 66.3. The improvement is especially large on TTL, increasing from 57.6 to 70.1. This suggests that ranking candidate memory transitions provides a stronger learning signal than directly using verifier scores as scalar rewards, leading to safer and more useful memory updates.

% \begin{figure}[!t]
% \centering
% \includegraphics[width=\columnwidth]{fig/2.2.png}
% % % % % % % \vspace{-5mm}
% \caption{
% Memory transition errors on the Mem-\(\alpha\) validation set. 
% We report the percentage of intermediate memory transitions containing omission, corruption, or hallucination. 
% Lower values indicate more reliability.
% }
% \label{fig:transition_level_errors}
% % % % % % % \vspace{-5mm}
% \end{figure}

% \subsection{Reliability of Memory Consolidation}
% \label{sec:memory_reliability}

% We further evaluate the reliability of intermediate memory transitions by measuring three semantic failure modes: omission, corruption, and hallucination. 
% We use generated trajectories from the Mem-\(\alpha\) validation set, covering 463 instances. 
% For each method, we construct chunk-level transition records and use GPT-4o-mini~\cite{hurst2024gpt} with temperature set to 0 as an LLM judge to identify whether each transition contains any of the three errors. 
% The judge follows the prompt and definitions provided in Appendix~\ref{app:transition_judge_prompt}. 
% We report the percentage of transitions containing each error type, counting only major errors that may affect later recall, classification, personalization, or QA.

\subsection{Reliability of Memory Consolidation}
\label{sec:memory_reliability}

We evaluate transition-level reliability by measuring three semantic failure modes: omission, corruption, and hallucination. 
Using generated trajectories from the 463-instance Mem-\(\alpha\) validation set, we construct chunk-level transition records for each method and use GPT-4o-mini~\cite{hurst2024gpt} with temperature 0 as an LLM judge. 
The judge follows the prompt and definitions in Appendix~\ref{app:transition_judge_prompt}; we report the percentage of transitions containing each major error type.

As shown in Figure~\ref{fig:transition_level_errors}, omission is the dominant error type because memory construction is inherently compressive, while corruption and hallucination require actively writing incorrect or unsupported facts. 
Thus, comparisons should be made within each error type across methods. 
TrustMem achieves the lowest error rates across all categories, with 8.08\% omission, 0.14\% corruption, and 0.01\% hallucination, demonstrating that transition-level verification reduces missing, distorted, and unsupported memory content.

\subsection{Training Reward Dynamics}
\label{sec:reward_dynamics}

\begin{figure}[t]
\centering
\includegraphics[width=\columnwidth]{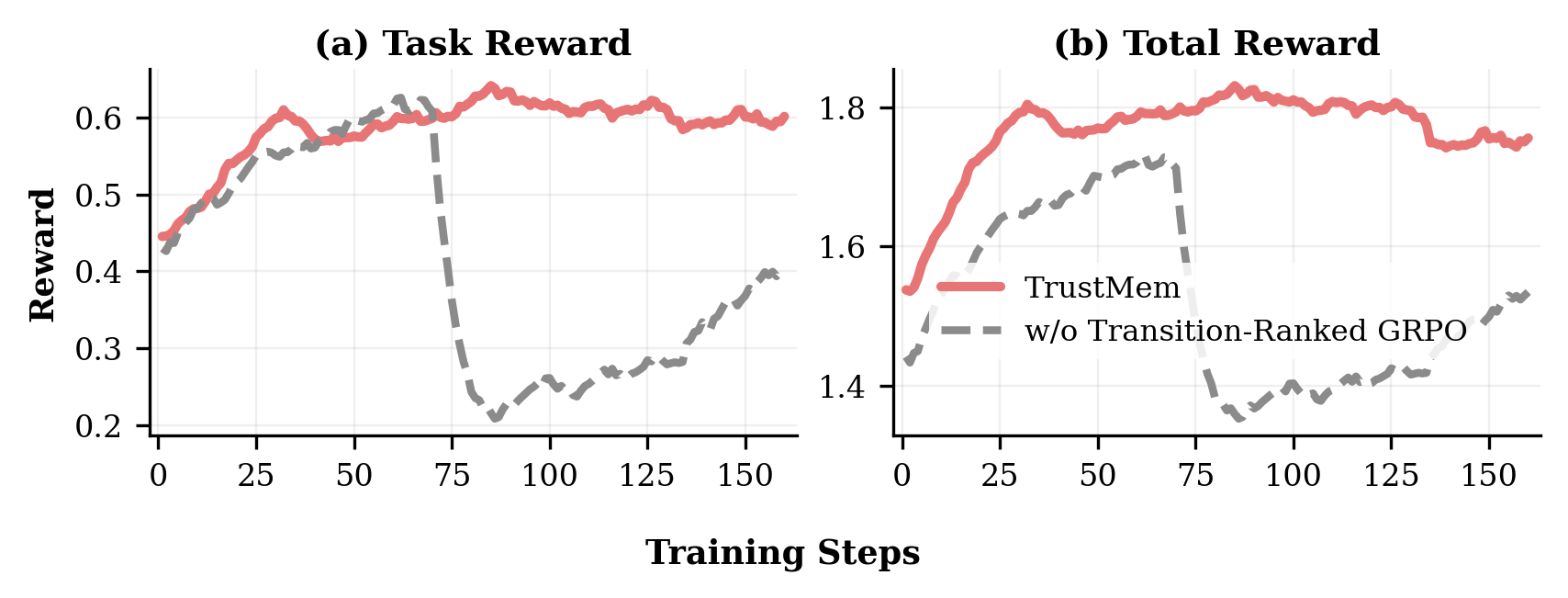}
% % % % % % \vspace{-5mm}
\caption{
Training reward dynamics of TrustMem and the variant without Transition-Ranked GRPO.
% Without transition-ranked optimization, the model shows unstable reward dynamics and a sharp drop after early improvement, whereas TrustMem maintains higher and more stable task and total rewards.
}
\label{fig:reward}
% % % % % % \vspace{-7mm}
\end{figure}

Figure~\ref{fig:reward} compares TrustMem with a variant trained without Transition-Ranked GRPO. 
Both methods improve in the early stage, suggesting that task-level rewards can learn basic memory-editing behaviors. 
However, the variant without transition-ranked optimization later becomes unstable and shows a sharp drop in both task and total rewards, while TrustMem maintains higher rewards and converges to a stable plateau. 
This indicates that coarse scalar rewards alone are insufficient for reliable memory consolidation. 
By ranking candidate memory transitions with verifier scores, Transition-Ranked GRPO provides finer-grained supervision over memory state changes, encouraging trustworthy updates and stabilizing training.

\subsection{Case Study}
\label{sec:case_study}

We present representative case studies in Figure~\ref{fig:intro_case}, with additional examples provided in Table~\ref{tab:memory_transition_case_studies} in the Appendix. These examples cover transition-level failures: corruption, omission, and hallucination.

Existing memory agents often produce unreliable memory updates. Mem1 reverses input semantics by storing a negation-reversed fact; AtomMem omits all key status-symbol facts; and MemAgent hallucinates user events from labeled command examples. Such errors are written into the memory and may affect future retrieval and reasoning.

In contrast, \textsc{TrustMem} preserves key evidence and avoids unsupported memory edits, demonstrating more faithful transition-level memory consolidation.

\input{tables/tab_5}

%% file: tables/tab_3.tex
\begin{table*}[!ht]
    \centering
    \small
    \resizebox{\textwidth}{!}{%
    \begin{tabular}{l|cccc|ccc|ccc}
        \toprule
        \multirow{2}{*}{\textbf{Method}}
        & \multicolumn{4}{c|}{\textbf{Memory Extraction}}
        & \multicolumn{3}{c|}{\textbf{Memory Updating}}
        & \multicolumn{3}{c}{\textbf{Question Answering}} \\
        \cmidrule(lr){2-5}
        \cmidrule(lr){6-8}
        \cmidrule(lr){9-11}
        &
        \textbf{R$\uparrow$}
        & \textbf{Weighted R$\uparrow$}
        & \textbf{Acc.$\uparrow$}
        & \textbf{F1$\uparrow$}
        & \textbf{C$\uparrow$}
        & \textbf{H$\downarrow$}
        & \textbf{O$\downarrow$}
        & \textbf{C$\uparrow$}
        & \textbf{H$\downarrow$}
        & \textbf{O$\downarrow$} \\
        \midrule

        Mem0~\citep{chhikara2025mem0}
        & \underline{42.91} & \underline{65.03} & 60.86 & \underline{57.31}
        & 25.50 & \underline{0.45} & 74.02
        & 53.02 & 19.17 & 27.81 \\

        Memobase~\citep{memobase2025}
        & 14.55 & 25.88 & 32.29 & 25.13
        & 5.20 & 0.55 & 94.25
        & 35.33 & 29.97 & 34.71 \\

        Supermemory~\citep{supermemory2025}
        & 41.53 & 64.76 & 60.83 & 56.90
        & 16.37 & 1.15 & 82.47
        & 54.07 & 22.24 & 23.69 \\

        Zep~\citep{rasmussen2025zep}
        & -- & -- & -- & --
        & \textbf{47.28} & \textbf{0.42} & 52.31
        & \underline{55.47} & 21.92 & 22.62 \\

        MemAgent~\citep{yu2025memagent}
        & 12.39 & 15.31 & \underline{82.02} & 22.03
        & 11.66 & 4.55 & 83.78
        & 42.11 & 28.70 & 28.12 \\

        Mem-T~\citep{yue2026mem}
        & 36.60 & 46.85 & 67.13 & 52.69
        & 45.53 & 3.35 & \underline{51.12}
        & \textbf{59.43} & 26.11 & \underline{13.98} \\

        AtomMem~\citep{huo2026atommem}
        & 0.28 & 0.39 & \textbf{87.50} & 0.56
        & 2.14 & 4.29 & 93.57
        & 31.27 & \textbf{5.56} & 64.00 \\

        \rowcolor{lightpurple}
        \ours
        & \textbf{56.73} & \textbf{67.45} & 67.81 & \textbf{69.45}
        & \underline{46.84} & 0.53 & \textbf{48.64}
        & 55.22 & \underline{6.07} & \textbf{12.46} \\

        \bottomrule
    \end{tabular}%
    }
    \vspace{-2mm}
    \caption{Experiment results on HaluMem Benchmark. We report operation-level metrics for memory extraction, memory updating, and memory-based question answering. R, Weighted R, Acc., and F1 evaluate memory extraction quality; C, H, and O denote correctness, hallucination, and omission, respectively.}
    \label{tab:halumem_medium_full_results}
    \vspace{-3mm}
\end{table*}

%% file: tables/tab_4.tex
\begin{table}[t]
    \centering
    \small
    \begin{tabular}{l|cccc}
    \toprule
    \textbf{Methods} 
    & \textbf{AR} 
    & \textbf{TTL} 
    & \textbf{LRU} 
    & \textbf{Overall} \\
    \midrule

    \rowcolor{tableHeaderBg}
    \multicolumn{5}{l}{\textit{Ablation of Reward Components}} \\
    w/o \(R_{\text{act}}\) 
        & 67.4 & 52.6 & 15.9 & 54.0 \\
    w/o \(R_{\text{eff}}\) 
        & 71.1 & 66.9 & 18.8 & 61.8 \\

    \midrule
    \rowcolor{tableHeaderBg}
    \multicolumn{5}{l}{\textit{Ablation of Transition Verification}} \\
    w/o \(R_{\text{ver}}\) 
        & 72.1 & 58.3 & 20.6 & 58.7 \\

    \midrule
    \rowcolor{tableHeaderBg}
    \multicolumn{5}{l}{\textit{Ablation of Optimization Strategy}} \\
    PRM-style GRPO 
        & 75.8 & 57.6 & 21.5 & 60.3 \\
    TrustMem (TR-GRPO) 
        & 76.4 & 70.1 & 22.6 & 66.3 \\
    \bottomrule
    \end{tabular}
    \vspace{-2mm}
    \caption{Ablation studies on reward components, transition verification, and optimization strategy.
    }
    \label{tab:ablation}
    \vspace{-5mm}
\end{table}

% AR, TTL, and LRU denote Accurate Retrieval, Test-Time Learning, and Long-Range Understanding, respectively.

%% file: tables/tab_5.tex
% \begin{table*}[t]
\begin{table*}[!t]
\centering
\scriptsize
\renewcommand{\arraystretch}{1.25}
\rowcolors{3}{tableRow}{white}
\begin{tabularx}{\textwidth}{p{0.32\textwidth} p{0.32\textwidth} X}
\toprule[1pt]
\rowcolor{tableHeaderBg}\textbf{Input Chunk} & \textbf{Existing Method Memory} & \textbf{\textsc{TrustMem} Memory} \\
\midrule

Paris became the capital of French literature in the 17th century
with Boileau, Corneille, La Fontaine, Moliere, and Racine.
\begingroup\sethlcolor{tableHeaderBg}\hl{In the 18th century, Parisian literary
life was dominated by Voltaire, Rousseau, Marivaux, and Beaumarchais.}\endgroup &
\textit{\textsc{Mem1}.}
\par\textbf{Stored memory:} Paris became a 17th-century literary center with
Boileau, Corneille, La Fontaine, Moliere, Racine, and
\begingroup\sethlcolor{red!15}\hl{Voltaire}\endgroup.
\par\textbf{\textcolor{red!70!black}{Corruption: wrong time period.}} &
\textbf{Ours.}
\par\textbf{Stored memory:} Paris literary timeline:
17C---Boileau, Corneille, La Fontaine, Moliere, Racine;
\begingroup\sethlcolor{green!15}\hl{18C---Voltaire, Rousseau, Marivaux,
Beaumarchais.}\endgroup \\

\addlinespace[3pt]

Constantine Monomachos restored the Church of the Holy Sepulchre between
1042 and 1048. \begingroup\sethlcolor{tableHeaderBg}\hl{Nothing survived of the
mosaics covering the walls and dome}\endgroup;
Russian abbot Daniel later left a description. &
\textit{\textsc{Mem1}.}
\par\textbf{Stored memory:} Daniel described mosaics as lively and
\begingroup\sethlcolor{red!15}\hl{covering the walls and dome}\endgroup.
\par\textbf{\textcolor{red!70!black}{Corruption: negation reversed.}} &
\textbf{Ours.}
\par\textbf{Stored memory:} Holy Sepulchre: restored by Constantine Monomachos
in 1042--1048; \begingroup\sethlcolor{green!15}\hl{wall/dome mosaics did not
survive}\endgroup; Daniel later described the decoration. \\

\addlinespace[3pt]

Emperor is a Norwegian black metal band formed in 1991.
\begingroup\sethlcolor{tableHeaderBg}\hl{It split in 2001; reunited in 2005--2007
and 2013--2014.}\endgroup It was founded by Ihsahn and Samoth. &
\textit{\textsc{Mem1}.}
\par\textbf{Stored memory:} Emperor is a Norwegian black metal band formed in
1991, founded by Ihsahn and Samoth.
\par\textbf{\textcolor{red!70!black}{Omission: missing split/reunion timeline.}} &
\textbf{Ours.}
\par\textbf{Stored memory:} Emperor: Norwegian black metal band, founded
1991 by Ihsahn/Samoth; \begingroup\sethlcolor{green!15}\hl{split 2001; reunions
in 2005--2007 and 2013--2014.}\endgroup \\

\addlinespace[3pt]

``Because the Night'' reached \#13 on Billboard,
\begingroup\sethlcolor{tableHeaderBg}\hl{number 5 in the UK}\endgroup,
and was ranked \begingroup\sethlcolor{tableHeaderBg}\hl{number 116}\endgroup on
\textit{NME}'s Top 150 Singles of All Time. &
\textit{\textsc{MemAgent}.}
\par\textbf{Stored memory:} ``Because the Night'' reached \#13 on Billboard.
\par\textbf{\textcolor{red!70!black}{Omission: missing UK peak and NME ranking.}} &
\textbf{Ours.}
\par\textbf{Stored memory:} ``Because the Night'': Billboard \#13;
\begingroup\sethlcolor{green!15}\hl{UK number 5; NME all-time rank number 116.}\endgroup \\

\addlinespace[3pt]

\begingroup\sethlcolor{tableHeaderBg}\hl{Only senators could wear Tyrian purple in
ancient Rome; high-ranking Hawaiian chiefs could wear feather cloaks; before the
republic, only the Chinese emperor could wear yellow.}\endgroup &
\textit{\textsc{AtomMem}.}
\par\textbf{Stored memory:}
\begingroup\sethlcolor{red!15}\hl{None.}\endgroup
\par\textbf{\textcolor{red!70!black}{Omission: all status-symbol facts omitted.}} &
\textbf{Ours.}
\par\textbf{Stored memory:} Clothing as status marker:
\begingroup\sethlcolor{green!15}\hl{Roman senators---Tyrian purple; Hawaiian
chiefs---feather cloaks; Chinese emperor---yellow.}\endgroup \\

\addlinespace[3pt]

Austrocylindropuntia is a cactus genus
\begingroup\sethlcolor{tableHeaderBg}\hl{with 11 recognized species. It was once
included in Opuntia; some species are native to South America.}\endgroup &
\textit{\textsc{MemT}.}
\par\textbf{Stored memory:} Austrocylindropuntia is a cactus genus \ldots
\begingroup\sethlcolor{red!15}\hl{John W. Ingram was president of the Rock
Island Railroad.}\endgroup
\par\textbf{\textcolor{red!70!black}{Hallucination: unsupported railroad fact.}} &
\textbf{Ours.}
\par\textbf{Stored memory:} Austrocylindropuntia:
\begingroup\sethlcolor{green!15}\hl{11-species cactus genus; formerly Opuntia;
some South American.}\endgroup \\

\addlinespace[3pt]

RAAF Base Edinburgh is a Royal Australian Air Force military airbase
\begingroup\sethlcolor{tableHeaderBg}\hl{located 25 km north of Adelaide, South
Australia.}\endgroup &
\textit{\textsc{MemT}.}
\par\textbf{Stored memory:} RAAF Base Edinburgh is a RAAF airbase \ldots
\begingroup\sethlcolor{red!15}\hl{EPMD is a hip-hop group; ``True North'' is a
TV episode; Bill Parcells coached the Giants.}\endgroup
\par\textbf{\textcolor{red!70!black}{Hallucination: unrelated pop-culture facts.}} &
\textbf{Ours.}
\par\textbf{Stored memory:} RAAF Base Edinburgh:
\begingroup\sethlcolor{green!15}\hl{RAAF military base near Edinburgh, South
Australia, 25 km north of Adelaide.}\endgroup \\

\addlinespace[3pt]

Input contains \begingroup\sethlcolor{tableHeaderBg}\hl{labeled command examples}\endgroup:
``\begingroup\sethlcolor{tableHeaderBg}\hl{add new email to contacts}\endgroup'';
``\begingroup\sethlcolor{tableHeaderBg}\hl{how is shell stock doing today}\endgroup'';
``\begingroup\sethlcolor{tableHeaderBg}\hl{confirm that the alarm is set for tomorrow}\endgroup''. &
\textit{\textsc{MemAgent}.}
\par\textbf{Stored memory:}
\begingroup\sethlcolor{red!15}\hl{John is the user's friend. At 2024-01-10, the
user added a new email to contacts and checked Shell stock.}\endgroup
\par\textbf{\textcolor{red!70!black}{Hallucination: examples treated as real events.}} &
\textbf{Ours.}
\par\textbf{Stored memory:} Command-intent examples:
\begingroup\sethlcolor{green!15}\hl{add email: label 1; check Shell stock:
label 4; confirm alarm: label 42.}\endgroup \\

\addlinespace[3pt]

\begingroup\sethlcolor{tableHeaderBg}\hl{Czech verb aspect pairs differ by prefix or
suffix; Czech has indicative, imperative, and conditional moods; Czech
orthography is phonemic and uses hacek marks.}\endgroup &
\textit{\textsc{AtomMem}.}
\par\textbf{Stored memory:} Czech verb aspects and moods \ldots
\begingroup\sethlcolor{red!15}\hl{the Digimon franchise, Digital World, and
Digidestined/Tamers.}\endgroup
\par\textbf{\textcolor{red!70!black}{Hallucination: unrelated franchise memory.}} &
\textbf{Ours.}
\par\textbf{Stored memory:} Czech grammar:
aspect via prefix/suffix; indicative/imperative/conditional moods;
\begingroup\sethlcolor{green!15}\hl{phonemic orthography with hacek marks.}\endgroup \\

\bottomrule[1pt]
\end{tabularx}
\caption{Representative memory transition case studies across multiple baseline agents. Each row shows the input chunk, the memory stored by an existing method, and the memory stored by \textsc{TrustMem}. Blue highlights mark relevant evidence in the input; red highlights mark corrupted or unsupported memory spans, and bold red annotations identify the failure type.}
\label{tab:memory_transition_case_studies}
\end{table*}

%% file: _sec_5_Challenges.tex
\section{Conclusion}

In this paper, we introduce \textsc{TrustMem}, a framework for learning trustworthy memory consolidation in LLM agents with long-term memory. Instead of relying only on final task outcomes, \textsc{TrustMem} supervises intermediate memory transitions through a Memory Transition Verifier and optimizes memory-editing behavior with Transition-Ranked GRPO. Experiments on MemoryAgentBench, HaluMem, and the Mem-\(\alpha\) show that \textsc{TrustMem} improves both memory utility and operation-level reliability. The results further demonstrate that transition-level verification helps reduce omission, corruption, and hallucination, highlighting its importance for reliable long-term memory systems.

\section*{Limitations}

A key limitation of \textsc{TrustMem} is that our current experiments mainly focus on text-based long-term memory benchmarks. 
However, in real-world agent applications, user inputs and memory evidence are often multimodal, involving not only textual interactions but also images, videos, documents, and other contextual signals. 
We plan to extend our framework to multimodal memory agents and vision-language models, enabling more faithful and trustworthy consolidation of multimodal memory for open-ended personalized applications.

%% file: _sec_6_Conclusion.tex
\clearpage
\appendix

\section{Experiment Settings}

\begin{table*}[t]
\centering
\small
\resizebox{\textwidth}{!}{
\begin{tabular}{l|l|l|l}
\toprule
\textbf{Method} & \textbf{Category} & \textbf{Model/System Used} & \textbf{Backbone/Base Model} \\
\midrule
Long-Context 
& Standard paradigm 
& Qwen3-32B with 32K context window
& Qwen3-32B \\

RAG-Top2 
& Standard paradigm 
& BM25 top-2 retrieval + Qwen3-32B
& Qwen3-32B answer model \\

MemAgent~\citep{yu2025memagent}
& RL-based / agentic memory method 
& RL-MemAgent-7B 
& Qwen2.5-7B \\

MEM1~\citep{zhou2025mem1}
& RL-based / agentic memory method 
& MEM1-7B 
& Qwen2.5-7B \\

Mem-\(\alpha\)~\citep{wang2025mem}
& RL-based memory method 
& Mem-\(\alpha\) 
& Qwen3-4B \\

Mem-T~\citep{yue2026mem}
& RL-based / agentic memory method 
& Mem-T-4B 
& Qwen3-4B \\

AtomMem~\citep{huo2026atommem}
& RL-based / agentic memory method 
& AtomMem-8B 
& Qwen3-8B \\

Mem0~\citep{chhikara2025mem0}
& System-level memory framework 
& Mem0 
& Not fixed \\

Memobase~\citep{memobase2025}
& System-level memory framework 
& Memobase 
& Not fixed \\

Supermemory~\citep{supermemory2025}
& System-level memory framework 
& Supermemory 
& Not fixed \\

Zep~\citep{rasmussen2025zep}
& System-level memory framework 
& Zep 
& Not fixed \\
\bottomrule
\end{tabular}
}
\caption{Summary of baseline models and systems used in our experiments. For model-based baselines, we report the evaluated checkpoint or model and its backbone when applicable. For system-level memory frameworks, the backbone is marked as ``Not fixed'' because these methods function as memory infrastructures and can be paired with different LLMs during extraction, updating, retrieval, or answer generation.}
\label{tab:baseline_model_details}
\end{table*}

The exact subset of baselines evaluated on each benchmark depends on the applicability of the corresponding method and the availability of comparable results. The reported baselines for each benchmark are shown in the corresponding result tables \ref{tab:baseline_model_details}.

\section{Experiments on Memory Utility}
\label{sec:exp_memory_utility_appendix}

Table~\ref{tab:main_result_mem-alpha} reports results on the validation set. \ours achieves the best average score of 66.3, outperforming Mem-\(\alpha\) by 4.4 points. The improvement is especially clear on Test-Time Learning, where \ours obtains 72.4 on TREC-C, 73.8 on NLU, and 64.0 on PubMed. \ours also improves BookSum from 18.7 to 22.6, showing stronger long-range memory utility. These results suggest that transition-level verification helps the agent preserve useful task-specific and long-range information during memory construction.

Table~\ref{tab:main_result_memoryagentbench} evaluates out-of-distribution generalization on MemoryAgentBench. \ours achieves the best average score of 65.7, improving over Mem-\(\alpha\) by 6.5 points and outperforming Long-Context and RAG-Top2 by 19.6 and 15.5 points, respectively. \ours performs strongly on both retrieval and test-time learning tasks, achieving 83.0 on Single-Doc, 85.0 on TREC-C, and 84.0 on Banking77. These results show that \ours learns a more robust memory update policy than simply retaining long context or retrieving top-ranked chunks. Overall, reliable local memory transitions improve both validation performance and out-of-distribution memory utility.

\section{Experiments on HaluMem}
\label{sec:exp_halumem_appendix}

Table~\ref{tab:halumem_medium_full_results} evaluates operation-level memory reliability on HaluMem-Medium, covering memory extraction, memory updating, and memory-based question answering. This benchmark directly matches the goal of \ours: improving the safety and reliability of each local memory transition.

For memory extraction, \ours achieves the best recall, weighted recall, and F1, with scores of 56.73, 67.45, and 69.45, respectively. Compared with the strongest prior F1 score from Mem0, \ours improves extraction F1 by 12.14 points. Although AtomMem obtains the highest extraction accuracy, its recall and F1 are extremely low, indicating that it misses most useful memories. In contrast, \ours achieves a better balance between coverage and correctness, showing that the verifier helps reduce both omission and unsupported memory writing.

For memory updating, \ours achieves the lowest omission rate of 48.64 and the second-best correctness score of 46.84. This supports the role of the Memory Transition Verifier, which checks whether each update preserves valid prior memory, incorporates new evidence, and avoids unsupported changes. As a result, \ours learns safer memory-editing behavior and reduces missed updates.

For memory-based question answering, \ours obtains the lowest omission rate of 12.46 and the second-lowest hallucination rate of 6.07 while maintaining competitive correctness. Compared with methods that either hallucinate more, such as Mem-T, or omit too much information, such as AtomMem, \ours provides a stronger trade-off across correctness, hallucination, and omission. Overall, these results show that transition-level verification and preference-guided optimization improve not only final task performance but also the intermediate memory operations required for trustworthy long-term memory.

\begin{figure*}[t]
\centering
\begin{tcolorbox}[
    enhanced,
    breakable,
    width=0.98\textwidth,
    colback=green!6,
    colframe=green!55!black,
    coltitle=white,
    colbacktitle=green!45!black,
    title={Prompt Template for Transition Verification},
    fonttitle=\bfseries,
    boxrule=0.8pt,
    arc=2mm,
    left=2mm,
    right=2mm,
    top=1mm,
    bottom=1mm
]
\small
\textbf{Role:} You are an expert judge for chunk-level memory consolidation.

\textbf{Goal:}
Given a compact view of one memory transition from \(M_i\) to \(M_{i+1}\) after processing a chunk \(c_i\), evaluate whether the memory update is trustworthy according to:
\begin{enumerate}
    \item \textbf{Coverage:} whether \(M_{i+1}\) preserves the write-worthy key information from \(c_i\);
    \item \textbf{Preservation:} whether important information already present in \(M_i\) is not deleted or overwritten without justification;
    \item \textbf{Faithfulness:} whether newly added or modified memory content is supported by the chunk evidence or the previous memory state.
\end{enumerate}

\textbf{Inputs:}
\begin{itemize}
    \item \textbf{Chunk Evidence:} \texttt{\{chunk\_text\}}
    \item \textbf{Pre-transition Memory State:} \texttt{\{pre\_memory\}}
    \item \textbf{Post-transition Memory State:} \texttt{\{post\_memory\}}
    \item \textbf{Touched Memory Actions:} \texttt{\{actions\}}
\end{itemize}

\textbf{Required Output:}
Return only a JSON object:
\begin{verbatim}
{
  "score": 0.0,
  "coverage_ok": true,
  "preservation_ok": true,
  "faithfulness_ok": true,
  "issues": ["brief issue labels"],
  "explanation": "brief explanation"
}
\end{verbatim}

\textbf{Rules:}
\begin{itemize}
    \item Judge only from the provided chunk evidence, pre-transition memory, post-transition memory, and touched memory actions.
    \item A skipped action can still be valid if the relevant information was already correctly stored and preserved.
    \item Legitimate merging, rewriting, or compression is allowed as long as important facts are preserved.
    \item Penalize critical omissions, unsupported additions, corrupted facts, and unjustified overwrites or deletions.
    \item Use a continuous score in \([0,1]\), where 1 means excellent consolidation and 0 means clearly invalid consolidation.
    \item Do not output extra text outside the required JSON object.
\end{itemize}
\end{tcolorbox}
\caption{Simplified prompt template for transition verification.}
\label{fig:transition-verification-prompt}
\end{figure*}

\clearpage
\onecolumn

\section{Memory Transition Error Judge Prompt}
\label{app:transition_judge_prompt}

We use GPT-4o-mini~\cite{hurst2024gpt} with temperature set to 0 as the LLM judge for transition-level memory reliability evaluation. 
The judge receives the prior memory, the current input chunk, the memory operations produced after reading the chunk, and the updated memory, and returns a structured JSON judgment over omission, corruption, and hallucination.

\paragraph{Evaluation Protocol.}
For the reliability analysis in Section~\ref{sec:memory_reliability}, we construct one record for each chunk-level memory transition. 
Each record contains the prior memory, the current input chunk, the generated memory-editing operations, and the updated memory. 
A transition may contain multiple error types, so we compute each error rate independently as the percentage of transitions marked with that error type. 
We count only errors marked as \texttt{major}; transitions with \texttt{minor} or \texttt{none} severity are not counted as failures.

\begin{tcolorbox}[
    enhanced,
    breakable,
    width=0.98\textwidth,
    colback=blue!5,
    colframe=blue!55!black,
    coltitle=white,
    colbacktitle=blue!45!black,
    title={Prompt Template for Memory Transition Error Judge},
    fonttitle=\bfseries,
    boxrule=0.8pt,
    arc=2mm,
    left=2mm,
    right=2mm,
    top=1mm,
    bottom=1mm
]
\small
\textbf{Role:} You are a strict but fair evaluator of memory-transition quality.

\textbf{Goal:}
Given one chunk-level memory transition, judge whether the transition introduces a substantive memory error. The judge evaluates three semantic failure modes:
\begin{enumerate}
    \item \textbf{Omission:} the updated memory misses key information from the input chunk.
    \item \textbf{Corruption:} the memory operation changes, contradicts, misattributes, or mixes a fact supported by the input chunk or prior memory.
    \item \textbf{Hallucination:} the memory operation introduces a substantive new fact unsupported by the input chunk or prior memory.
\end{enumerate}

\textbf{Inputs:}
\begin{itemize}
    \item \textbf{Dataset Group:} \texttt{\{dataset\_group\}}
    \item \textbf{Sample ID:} \texttt{\{sample\_id\}}
    \item \textbf{Chunk Index:} \texttt{\{chunk\_idx\}}
    \item \textbf{Prior Memory:} \texttt{\{prior\_memory\}}
    \item \textbf{Input Chunk:} \texttt{\{input\_chunk\}}
    \item \textbf{Memory Operations:} \texttt{\{memory\_operations\}}
\end{itemize}

\textbf{Definitions:}
\begin{itemize}
    \item \textbf{Key information} includes facts, labels, constraints, preferences, entities, relations, dates, numbers, or outcomes needed for later question answering, classification, personalization, or faithful recall.
    \item A concise summary is allowed; skipping nonessential details alone is not counted as omission.
    \item Facts already supported by prior memory are not counted as hallucinations when they are simply preserved by an update.
\end{itemize}

\textbf{Required Output:}
Return only a JSON object:
\begin{verbatim}
{
  "omission": true,
  "corruption": false,
  "hallucination": false,
  "primary_error": "omission",
  "severity": "major",
  "confidence": 85,
  "rationale": "brief explanation",
  "evidence": ["supporting evidence strings"]
}
\end{verbatim}

\textbf{Rules:}
\begin{itemize}
    \item Judge only from the provided prior memory, input chunk, and memory operations.
    \item Do not penalize harmless meta text, timestamps, assistant acknowledgements, or memory bookkeeping unless it changes a substantive fact.
    \item For classification/example chunks, labels and representative examples are important.
    \item For long document chunks, central entities, events, relations, and numeric/date facts are important; minor wording differences are not errors.
    \item For omission, ask whether a future assistant would likely answer incorrectly or fail the benchmark because the key information was not stored.
    \item If there are multiple major error types, set \texttt{primary\_error} to \texttt{"mixed"}.
    \item Use \texttt{"major"} only when the error would plausibly harm later recall or QA; use \texttt{"minor"} for small issues that probably would not matter.
    \item Use \texttt{"none"} when no substantive error is present.
    \item Do not output extra text outside the required JSON object.
\end{itemize}
\end{tcolorbox}

\twocolumn